\documentclass[a4paper]{article}
\usepackage{spconf,amsmath,graphicx}
\usepackage{amsfonts}
\usepackage{amssymb}
\usepackage{multirow}
\usepackage{bm}
\usepackage{subfigure}
\usepackage{url}
\usepackage{multirow}
\usepackage{algorithm}
\usepackage{algorithmicx}
\usepackage{algpseudocode}
\usepackage{lipsum}
\usepackage{multicol}

\title{Variational Regularized Transmission Refinement for Image Dehazing}
\name{Qiaoling Shu$^{\dag,\ddag}$, Chuansheng Wu$^{\dag}$, Zhe Xiao$^{\S}$, and Ryan Wen Liu$^{\ddag,\star}$\thanks{This work was supported by NSFC (No.: 51609195).}} 
\address{$^{\dag}$Department of Mathematics, $^{\ddag}$School of Navigation, Wuhan University of Technology, China\\$^{\S}$Institute of High Performance Computing, A*STAR, Singapore\\$^{\star}$Email: wenliu@whut.edu.cn}
\begin{document}
%
\maketitle
\begin{abstract}
High-quality dehazing performance is highly dependent upon the accurate estimation of transmission map. In this work, the coarse estimation version is first obtained by weightedly fusing two different transmission maps, which are generated from foreground and sky regions, respectively. A hybrid variational model with promoted regularization terms is then proposed to assisting in refining transmission map. The resulting complicated optimization problem is effectively solved via an alternating direction algorithm. The final haze-free image can be effectively obtained according to the refined transmission map and atmospheric scattering model. Our dehazing framework has the capacity of preserving important image details while suppressing undesirable artifacts, even for hazy images with large sky regions. Experiments on both synthetic and realistic images have illustrated that the proposed method is competitive with or even outperforms the state-of-the-art dehazing techniques under different imaging conditions.
\end{abstract}
\section{Introduction}
\label{sec:intro}
The presence of haze or fog can significantly degrade the visibility of an image captured in outdoor environments. Recovering high-quality images from degraded images (a.k.a. image dehazing) is beneficial for many realistic applications, e.g. video surveillance, unmanned vehicles, object recognition and tracking, etc \cite{SinghACME2018}. The atmospheric scattering model can be used to describe the hazy image generation process, i.e.,
\begin{equation}\label{Eq:PhysicalModelV1}
   \mathbf{I} \left( \mathrm{x} \right) = \mathbf{J} \left( \mathrm{x} \right) t \left( \mathrm{x} \right) + \left( 1 - t \left( \mathrm{x} \right) \right) \mathbf{A},
\end{equation}
where $\mathbf{I}$ is the observed hazy image, $\mathbf{J}$ is the haze-free scene radiance to be restored, $\mathbf{A}$ is the global atmospheric light, and $t$ is the transmission map related to depth map. The purpose of image dehazing is to recover $\mathbf{J}$ from $\mathbf{I}$, which is particularly challenging since both transmission $t$ and atmospheric light $\mathbf{A}$ are unknown. Several physically grounded priors, e.g., dark channel prior (DCP) \cite{HePAMI2011}, color-lines prior \cite{FattalTOG2014}, color attenuation prior \cite{ZhuMaiTIP2015}, non-local prior \cite{BermanCVPR2016}, and color ellipsoid prior \cite{BuiKimTIP2018}, have been proposed to assist in improving image dehazing. We will mainly consider the DCP-based dehazing methods since other priors fall beyond the focus of this work.

It is well known that high-quality dehazing performance is strongly dependent on the accurate estimation of transmission map. Many efforts have been devoted to refine the DCP-based coarse transmission map, such as soft matting method \cite{HePAMI2011}, guided image filtering \cite{HeSunPAMI2013}, total generalized variation (TGV) \cite{ShuICONIP2018}, image guided TGV \cite{ChenECCV2016}, non-local total variation \cite{LiuGaoTIP2018}, and kernel regression model \cite{XieSIVP2017}, etc. Simultaneous estimation of transmission map and haze-free image have also been performed to enhance image dehazing \cite{LiuLiICASSP2018}. Note that transmission map is inversely proportional to depth map. Joint variational regularized methods \cite{FangSIAM2014,LiuICASSP2018} were thus presented to implement simultaneous depth map estimation and sharp image restoration. Traditional DCP easily generates block artifacts in estimated transmission map leading to image quality degradation. Therefore, several extensions of DCP, including multiscale opening DCP \cite{LiuAccess2017}, saliency-based DCP \cite{ZhangIETIP2018}, and sphere-guided DCP \cite{LiHuGRSL2018}, etc, have recently been developed to overcome the limitation. If an image contains substantial sky regions, DCP-based methods easily fail since DCP assumption is based on statistical analysis in non-sky regions.

With the rapid developments in deep learning, the popular convolutional neural network (CNN) and its generations have received remarkable dehazing results. For example, DehazeNet \cite{CaiXuTIP2016} and its multi-scale version \cite{RenECCV2016} were trained to estimate the transmission map. AOD-Net \cite{LiPengICCV2017} and FEED-Net \cite{ZhangRenICME2018} directly restored the latent sharp image from a hazy image through a light-weight CNN. Li \textit{et al.} \cite{LiGuoACCESS2018} proposed a flexible cascaded CNN which jointly estimated transmission map and atmospheric light by separate CNNs. Proximal Dehaze-Net \cite{YangSunECCV2018} was recently presented by incorporating the haze imaging model, dark channel and transmission priors into a deep architecture. Learning-based dehazing performance is essentially dependent upon the diversity and volume of training datasets. It is difficult to guarantee high-quality dehazing results under some imaging conditions. Please refer to recent reviews \cite{SinghACME2018,LiRenTIP2019} for more progresses on image dehazing.

To make dehazing more stable and flexible, a traditional but effective two-step transmission map estimation is proposed in this work. The coarse transmission map in the first step is generated by weightedly summing up two different transmissions, respectively, estimated from foreground and sky regions. A joint variational regularized model with hybrid constraints is then proposed in the second step to refine the coarse transmission map. The final sharp image can be directly obtained through the atmospheric scattering model.
\begin{figure}[t]
	\centering
	\includegraphics[width=\linewidth]{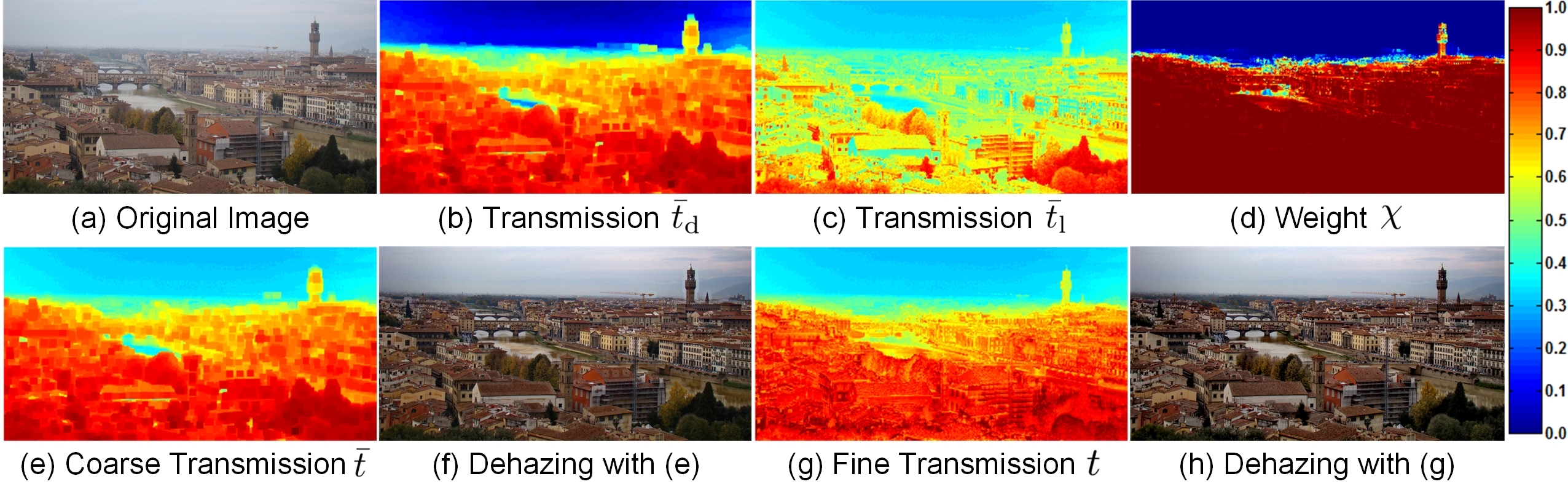}
	\caption{Our dehazing framework. From top-left to bottom-right: (a) input hazy image, (b) DCP-based transmission map $\bar{t}_{\mathrm{d}}$, (c) luminance-based transmission map $\bar{t}_{\mathrm{l}}$, (d) weight $\chi$ between $\bar{t}_{\mathrm{d}}$ and $\bar{t}_{\mathrm{l}}$, (e) weighted fusion-based coarse transmission map $\bar{t}$, (f) dehazing result with $\bar{t}$ in Fig. \ref{Figure0}(e), (g) refined transmission map $t$ and (h) dehazing result with $t$ in Fig. \ref{Figure0}(g).} 
	\label{Figure0}
\end{figure}
\section{Weighted Fusion-Based Coarse Transmission Map Estimation}
\label{sec:problem}
Natural images are commonly composed of foreground and sky (i.e., background) regions. The transmission map in foreground regions is directly estimated based on DCP assumption \cite{HePAMI2011} in this work. However, DCP commonly fails when there exists large sky regions. The luminance model \cite{ZhuTangNeuro2018} can be adopted to assist in estimating transmission map in sky regions. In particular, DCP-based transmission map $\bar{t}_{\mathrm{d}}$ is defined as $\bar{t}_{\mathrm{d}} \left( \mathrm{x} \right) = 1 - \omega \min_{c \in \left\{ r,g,b \right\}} \left( \min_{\mathrm{y} \in \Omega \left( \mathrm{x} \right)}  \left( \frac{\mathbf{I}_{c}(\mathrm{x})}{\mathbf{A}_{c}} \right) \right)$ with a control parameter $\omega = 0.95$ and a $21 \times 21$ region $\Omega \left( \mathrm{x} \right)$ centered at $\mathrm{x}$. The luminance-based transmission map $\bar{t}_{\mathrm{l}}$ is given by $\bar{t}_{\mathrm{l}} \left( \mathrm{x} \right) = \exp(- \beta \hat{L} \left( \mathrm{x} \right) )$ with $\beta$ being the scattering coefficient and $\hat{L}$ denoting the modified luminance value. From the optical point of view, $\beta$ is strongly correlated with wavelength. For red, green and blue channels in color images, the related coefficients $\beta$ are, respectively, selected as $0.3324$, $0.3433$ and $0.3502$ in our experiments. To reliably describe the influence of depth map on transmission map, the modified luminance $\hat{L}$ is corrected as follows
\begin{equation}\label{Eq:Luminance}
   \hat{L} \left( \mathrm{x} \right) = \frac{\tau}{L^*} L \left( \mathrm{x} \right),
\end{equation}
where $L$ is the luminance of an input image $\mathbf{I}$, $\tau$ describes the depth range\footnote{This parameter should be optimized to guarantee that there is a similar distribution range between DCP- and luminance-based transmission maps.}, $L^*$ denotes the $95\%$ percentile value of the luminance $L$. The final coarse transmission map $\bar{t}$ is obtained by weightedly fusing $\bar{t}_{\mathrm{d}}$ and $\bar{t}_{\mathrm{l}}$, i.e.,
\begin{equation}\label{Eq:CoarseTransmission}
   \bar{t}\left( \mathrm{x} \right) = \chi\left( \mathrm{x} \right) \bar{t}_{\mathrm{d}}\left( \mathrm{x} \right) + \left( 1 - \chi\left( \mathrm{x} \right) \right) \bar{t}_{\mathrm{l}}\left( \mathrm{x} \right),
\end{equation}
where the transmission weight $\chi \in [0, 1]$. If one pixel $\mathrm{x} \in \Omega$ belongs to the foreground regions, $\chi\left( \mathrm{x} \right)$ will tend to $1$ and $\bar{t}\left( \mathrm{x} \right) \rightarrow \bar{t}_{\mathrm{d}}\left( \mathrm{x} \right)$; conversely, $\chi\left( \mathrm{x} \right)$ will tend to $0$ and $\bar{t}\left( \mathrm{x} \right) \rightarrow \bar{t}_{\mathrm{l}}\left( \mathrm{x} \right)$. It is well known that DCP-based transmission map is uniformly small in sky regions. In contrast, the larger values could be found in foreground regions. As done in Ref. \cite{ZhuTangNeuro2018}, the weight function $\chi$ is given by $\chi\left( \mathrm{x} \right) = \frac{1}{1 + e^{-\theta_1 \bar{t}_{\mathrm{d}}\left( \mathrm{x} \right) - \theta_2}}$ with $\theta_1 = \frac{20}{\max(\bar{t}_{\mathrm{d}}) - \min(\bar{t}_{\mathrm{d}})}$ and $\theta_2 = -10 - \theta_1 \times \min(\bar{t}_{\mathrm{d}})$. Please refer to \cite{ZhuTangNeuro2018} for more details on luminance-based transmission map. As shown in Fig. \ref{Figure0}, it is difficult to yield high-quality dehazing results by directly adopting the coarse transmission map $\bar{t}$. We will propose a variational regularized model with hybrid constraints to further refine coarse transmission map.
\section{Fine Transmission Map Estimation with Promoted Regularization}
\label{sec:proposedmodel}
Before proposing our transmission map refinement method, we tend to deduce a more compact imaging model which rewrites the original image formulation model (\ref{Eq:PhysicalModelV1}) as follows
\begin{equation}\label{Eq:PhysicalModelV2}
   \bar{\mathbf{I}} \left( \mathrm{x} \right) = \bar{\mathbf{J}} \left( \mathrm{x} \right) t \left( \mathrm{x} \right),
\end{equation}
with $\bar{\mathbf{I}} \left( \mathrm{x} \right) = \mathbf{A} - \mathbf{I} \left( \mathrm{x} \right)$ and $\bar{\mathbf{J}} \left( \mathrm{x} \right) = \mathbf{A} - \mathbf{J} \left( \mathrm{x} \right)$ for $\mathrm{x} \in \Omega$. To guarantee robust estimation of transmission map, the initial estimation of $\bar{\mathbf{J}}$ in our experiments is given by $\bar{\mathbf{J}}^{0} \left( \mathrm{x} \right) = \frac{\mathbf{A} - \mathbf{I} \left( \mathrm{x} \right)}{\max \left( \bar{t} \left( \mathrm{x} \right), t_{\varepsilon} \right)}$, 
%
%
where $t_{\varepsilon}$ is a small constant to prevent imaging instability. For the sake of simplicity, we directly select ${I} = \mathbf{I}_{c}$, $\bar{I} = \bar{\mathbf{I}}_{c}$ and $\bar{J} = \bar{\mathbf{J}}_{c}$ for $c \in \left\{ r,g,b \right\}$. To enhance dehazing performance, the variational model with hybrid regularization terms for transmission map refinement is proposed as follows
\begin{align}\label{Eq:JointModel}
   \min_{\bar{J}, t} \Big\{ & \frac{\lambda_{1}}{2} \left\| \bar{I} - \bar{J} t \right\|_{2}^{2} + \frac{\lambda_{2}}{2} \left\| t - \bar{t} \right\|_{2}^{2} \\
   & + \lambda_{3} \left\| W \circ \left( \nabla t - \nabla I \right) \right\|_{1} + \lambda_{4} \left\| \nabla \bar{J} \right\|_{1} + \lambda_{5} \left\| \nabla t \right\|_{1} \Big\}, \nonumber
\end{align}
where $\lambda_{1 \leq i \leq 5}$ is a positive regularization parameter. In Eq. (\ref{Eq:JointModel}), the first two terms can be considered as a squared L2-norm data-fidelity term. The third L1-norm regularization is used to preserve the edges of transmission map. The last two terms are total variation (TV) regularizers which can stabilize the estimation process. The weighting function $W$ is selected as $W = e^{- \gamma \left\| \nabla I \right\|_{2}^{2} }$ with $\gamma$ being a control parameter. It is able to distinguish the edge and homogeneous regions. The proposed model (\ref{Eq:JointModel}) is thus capable of preserving the edges while suppressing the unwanted artifacts in homogeneous regions.

Due to the nonsmooth L1-norm penalties in Eq. (\ref{Eq:JointModel}), it is computationally intractable to generate stable solutions through traditional numerical methods. This paper proposes to develop an alternating direction algorithm to effectively handle the nonsmooth optimization problem (\ref{Eq:JointModel}). We first introduce three intermediate variables $X = \nabla t - \nabla I$, $Y = \nabla \bar{J}$ and $Z = \nabla t$, and then transform the unconstrained optimization problem (\ref{Eq:JointModel}) into the following constrained version
\begin{align}\label{Eq:constrainedproblem}
   &\min_{X,Y,Z,\bar{J},t} \Big\{ \frac{\lambda_{1}}{2} \left\| \bar{I} - \bar{J} t \right\|_{2}^{2} + \frac{\lambda_{2}}{2} \left\| t - \bar{t} \right\|_{2}^{2} \nonumber \\
   &~~~~~~~~~~~~~~~~~~~~~~~~~~~~~~ + \lambda_{3} \left\| W \circ X \right\|_{1} + \lambda_{4} \left\| Y \right\|_{1} + \lambda_{5} \left\| Z \right\|_{1} \Big\} \nonumber\\
   &~~~~~~\mathrm{s.t.}~~~~~X = \nabla t - \nabla I,~Y = \nabla \bar{J},~Z = \nabla t,
\end{align}
whose augmented Lagrangian function can be formulated as $\mathcal{L}_{\mathcal{A}} = \frac{\lambda_{1}}{2} \big\| \bar{I} - \bar{J} t \big\|_{2}^{2} + \frac{\lambda_{2}}{2} \big\| t - \bar{t} \big\|_{2}^{2} + \lambda_{3} \big\| W \circ X \big\|_{1} + \lambda_{4} \big\| Y \big\|_{1} + \lambda_{5} \big\| Z \big\|_{1} + \frac{\beta_{1}}{2} \big\| X - \big( \nabla t - \nabla I \big) - \frac{\xi}{\beta_{1}} \big\|_{2}^{2} + \frac{\beta_{2}}{2} \big\| Y - \nabla \bar{J} - \frac{\eta}{\beta_{2}} \big\|_{2}^{2} + \frac{\beta_{3}}{2} \big\| Z - \nabla t - \frac{\zeta}{\beta_{3}} \big\|_{2}^{2}$, where $\xi$, $\eta$ and $\zeta$ denote the Lagrangian multipliers, $\beta_{1}$, $\beta_{2}$ and $\beta_{3}$ are predefined positive parameters. The alternating direction method of multipliers (ADMM) is adopted to decompose $\mathcal{L}_{\mathcal{A}}$ into several subproblems with respect to $X$, $Y$, $Z$, $\bar{J}$ and $t$. We now alternatively solve these subproblems until the solution converges to the optimal value.

\textbf{$\left(X, Y, Z\right)$-subproblems}: Given the fixed values of $\bar{J}$ and $t$, $\left(X, Y, Z\right)$-subproblems are essentially L1-regularized least-squares programs, i.e.,
\begin{equation*}\label{Eq:subproblemX}
   X \leftarrow \min_{X} \Big\{ \lambda_{3} \big\| W \circ X \big\|_{1} + \frac{\beta_{1}}{2} \big\| X - \big( \nabla t - \nabla I \big) - \frac{\xi}{\beta_{1}} \big\|_{2}^{2} \Big\},
\end{equation*}
\begin{equation*}\label{Eq:subproblemY}
   Y \leftarrow \min_{Y} \Big\{ \lambda_{4} \big\| Y \big\|_{1} + \frac{\beta_{2}}{2} \big\| Y - \nabla \bar{J} - \frac{\eta}{\beta_{2}} \big\|_{2}^{2} \Big\},
\end{equation*}
\begin{equation*}\label{Eq:subproblemZ}
   Z \leftarrow \min_{Z} \Big\{ \lambda_{5} \big\| Z \big\|_{1} + \frac{\beta_{3}}{2} \big\| Z - \nabla t - \frac{\zeta}{\beta_{3}} \big\|_{2}^{2} \Big\},
\end{equation*}
whose solutions can be obtained using the following shrinkage operator \cite{BeckTeboulleSIAM2009}, i.e.,
\begin{equation}\label{Eq:subproblemXsolution}
   X \leftarrow {\mathbf{shrinkage}}\left( \nabla t - \nabla I + {\xi}/{\beta_{1}}, {\lambda_{3} W}/{\beta_{1}} \right),
\end{equation}
\begin{equation}\label{Eq:subproblemYsolution}
   Y \leftarrow {\mathbf{shrinkage}}\left( \nabla \bar{J} + {\eta}/{\beta_{2}}, {\lambda_{4}}/{\beta_{2}} \right),
\end{equation}
\begin{equation}\label{Eq:subproblemZsolution}
   Z \leftarrow {\mathbf{shrinkage}}\left( \nabla t + {\zeta}/{\beta_{3}}, {\lambda_{5}}/{\beta_{3}} \right),
\end{equation}
where the shrinkage operator is $\mathbf{shrinkage} \left(a, b \right) = \max (|a| - b,0) \circ \mathbf{sign} \left( a \right)$ with $\mathbf{sign}$ denoting the signum function.

\textbf{$\left(\bar{J}, t\right)$-subproblems}: Given the fixed values of $X$, $Y$ and $Z$ obtained from previous iterations, the minimizations of $\mathcal{L}_{\mathcal{A}}$ with respect to $\bar{J}$ and $t$ are equivalent to solving the following least-squares optimization problems
\begin{equation*}
\small
\begin{cases}
\begin{split}
& \bar{J} \leftarrow \min_{\bar{J}} \Big\{ \frac{\lambda_{1}}{2} \big\| \bar{I} - \bar{J} t \big\|_{2}^{2} + \frac{\beta_{2}}{2} \big\| Y - \nabla \bar{J} - \frac{\eta}{\beta_{2}} \big\|_{2}^{2} \Big\}\\
& t \leftarrow \min_{t} \Big\{ \frac{\lambda_{1}}{2} \big\| \bar{J} t - \bar{I} \big\|_{2}^{2} + \frac{\lambda_{2}}{2} \big\| t - \bar{t} \big\|_{2}^{2} + \frac{\beta_{1}+\beta_{3}}{2} \big\| \nabla t - \psi \big\|_{2}^{2} \Big\}
\end{split}
\end{cases}
\label{eq:subproblemJt}
\end{equation*}
%
%
where $\psi = \frac{\beta_{1}\hat{X} + \beta_{3}\hat{Z}}{\beta_{1}+\beta_{3}}$ with $\hat{X} = X + \nabla I - \frac{\xi}{\beta_{1}}$ and $\hat{Z} = Z - \frac{\zeta}{\beta_{3}}$. Let $\mathcal{F}$ be the forward fast Fourier transform (FFT) operator. The closed-form solutions $\bar{J}$ and $t$ can be directly obtained using the forward and inverse FFT operators, i.e.,
\begin{equation}\label{Eq:subproblemJsolution}
\footnotesize
   \bar{J} \leftarrow \mathcal{F}^{-1} \left(  \frac{ \lambda_{1} \mathcal{F} \left( \bar{I} / t \right) + \beta_{2} \overline{\mathcal{F} \left( \nabla \right)} \mathcal{F} \left( Y - {\eta}/{\beta_{2}} \right)}{\lambda_{1} \mathcal{F} \left( \textbf{I} \right) + \beta_{2} \overline{\mathcal{F} \left( \nabla \right)} \mathcal{F} \left( \nabla \right) } \right),
\end{equation}
\begin{equation}\label{Eq:subproblemtsolution}
\footnotesize
   t \leftarrow \mathcal{F}^{-1} \left(  \frac{ \lambda_{1} \mathcal{F} \left( \bar{I} / \bar{J} \right) + \lambda_{2} \mathcal{F} \left( \bar{t} \right) + \left( \beta_{1} + \beta_{3} \right) \overline{\mathcal{F} \left( \nabla \right)} \mathcal{F} \left( \psi \right)}{ \left( \lambda_{1} + \lambda_{2} \right) \mathcal{F} \left( \textbf{I} \right) + \left( \beta_{1} + \beta_{3} \right) \overline{\mathcal{F} \left( \nabla \right)} \mathcal{F} \left( \nabla \right) } \right),
\end{equation}
where $\textbf{I}$ is an identity matrix, $\mathcal{F}^{-1} \left( \cdot \right)$ is the inverse FFT operator, and $\overline{\mathcal{F} \left( \cdot \right)}$ denotes the complex conjugate operator.

\textbf{$\xi$, $\eta$ and $\zeta$ update}: During each iteration, the Lagrangian multipliers $\xi$, $\eta$ and $\zeta$ can be easily updated using $\xi \leftarrow \xi - \upsilon \beta_{1} \left( X - \left( \nabla t - \nabla I \right) \right)$, $\eta \leftarrow \eta - \upsilon \beta_{2} \left( Y - \nabla \bar{J} \right)$ and $\zeta \leftarrow \zeta - \upsilon \beta_{3} \left( Z - \nabla t \right)$ with $\upsilon$ being a steplength.

Note that the estimation of $\bar{J}$ from Eq. (\ref{Eq:subproblemJsolution}) easily suffers from the loss of fine textures. In this work, we still propose to restore the latent haze-free image $\mathbf{J}$ based on the estimated transmission map $t$ in Eq. (\ref{Eq:subproblemtsolution}). According to the image formulation model (\ref{Eq:PhysicalModelV1}), the final haze-free image $\mathbf{J}$ is given by
\begin{equation}\label{Eq:FinalImage}
   \mathbf{J} \left( \mathrm{x} \right) = \frac{\mathbf{I} \left( \mathrm{x} \right) - \mathbf{A}}{\max \left( t \left( \mathrm{x} \right), t_{\varepsilon} \right)} + \mathbf{A},
\end{equation}
which has been proven effective in current literature. 
%
%
%
\section{Experimental Results and Discussion}
Comprehensive experiments were performed using MATLAB R2017a on a machine with a 3.00 GHz Intel(R) Core (TM) i5-8500 CPU and 8.00 GB RAM. Both synthetic and realistic images were selected to compare our proposed method with several state-of-the-art dehazing methods, e.g., He-13 \cite{HeSunPAMI2013}, Ren-16 \cite{RenECCV2016}, Chen-16 \cite{ChenECCV2016}, Berman-16 \cite{BermanCVPR2016} and Liu-17 \cite{LiuCVIU2017}. In all experiments, the optimal parameters were manually selected for our proposed method, i.e., $\tau=3.4$, $\lambda_{1} = 1 \times 10^{-2}$, $\lambda_{2} = 5 \times 10^{-1}$, $\lambda_{3} = 5$, $\lambda_{4} = \lambda_{5} = 1$, $\beta_{1} = \beta_{2} =\beta_{3} = 1$, $t_{\varepsilon} = 1 \times 10^{-1}$, $\gamma = 2 \times 10^{2}$ and $\upsilon = \frac{1 + \sqrt{5}}{2}$. The effectiveness of these manually defined parameters for our method has been demonstrated through numerous experiments. To make fair comparisons, other competing dehazing methods were implemented with the best tuning parameters. Our Matlab source code is available at \url{http://mipc.whut.edu.cn}.
\setlength{\tabcolsep}{0.6pt}
\begin{table}[t]
	\footnotesize
	\centering
	\caption{Comparisons of PSNR/SSIM results for competing dehazing methods on different synthetic images.} 
	\begin{tabular}{|l|c|c|c|c|c|}
		\hline
		{Methods} & {Image 1} & {Image 2} & {Image 3} & {Image 4} & {Image 5} \\
		\hline \hline
		He-13 \cite{HeSunPAMI2013}         & 17.85/0.767 & 18.01/0.917 & 22.24/0.941 & \textbf{24.44}/\textbf{0.962} & 21.72/0.928\\ \hline
		Ren-16 \cite{RenECCV2016}       & 16.14/0.799 & 16.13/0.814 & 17.11/0.838 & 16.59/0.796 & 16.82/0.818\\ \hline
		Chen-16 \cite{ChenECCV2016}     & 19.32/0.838 & 18.43/0.875 & 24.16/0.908 & 14.35/0.610 & 21.72/0.910\\ \hline
		Berman-16 \cite{BermanCVPR2016} & 18.46/0.804 & 20.82/0.877 & 21.00/0.901 & 18.91/0.812 & 22.76/0.882\\ \hline
		Liu-17 \cite{LiuCVIU2017}       & 19.03/0.807 & 19.98/0.907 & 23.42/0.925 & 20.10/0.902 & 19.94/0.903\\ \hline
		Ours                            & \textbf{21.38}/\textbf{0.869} & \textbf{22.16}/\textbf{0.938} & \textbf{26.23}/\textbf{0.946} & 20.74/0.940 & \textbf{23.22}/\textbf{0.931}\\
		\hline
	\end{tabular}\label{TablePSNRSSIM}
\end{table}
\begin{figure}[t]
	\centering
	\includegraphics[width=\linewidth]{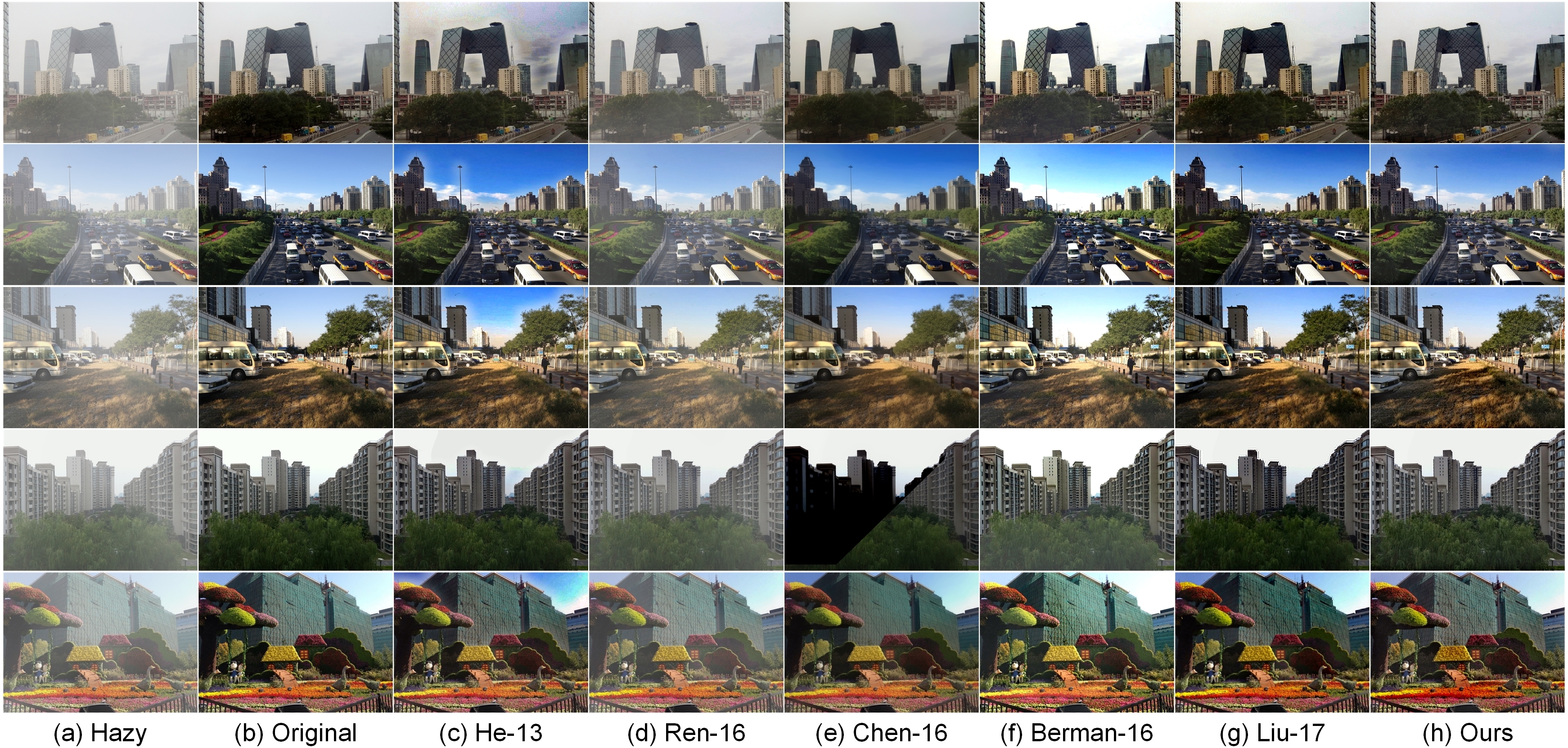}
	\caption{Comparisons of dehazing results on five different synthetic degraded images from \cite{LiRenTIP2019}. From left to right: (a) hazy image, (b) original image, dehazed images generated by (c) He-13 \cite{HeSunPAMI2013}, (d) Ren-16 \cite{RenECCV2016}, (e) Chen-16 \cite{ChenECCV2016}, (f) Berman-16 \cite{BermanCVPR2016}, (g) Liu-17 \cite{LiuCVIU2017} and (h) ours.} 
	\label{Figure1}
\end{figure}
\subsection{Experiments on Synthetic Images}
Synthetic experiments were implemented on five pairs of hazy and sharp images with sky regions, which were manually selected from the newly-released benchmark \cite{LiRenTIP2019}. Both PSNR and SSIM were adopted to quantitatively evaluate the dehazing results. Table \ref{TablePSNRSSIM} detailedly depicts the quantitative results for six different dehazing methods. It can be found that our method generates the superior imaging results under consideration in most of the cases. The deep learning-based dehazing method \cite{RenECCV2016} easily suffers from the lowest values of PSNR and SSIM. That may be due to the fact that learning-based dehazing methods are often sensitive to the volume and diversity of training datasets. He-13 \cite{HeSunPAMI2013} sometimes yields the highest quantitative results but easily brings halo and color aliasing artifacts in restored images. The visual results illustrated in Fig. \ref{Figure1} have further confirmed the advantages of our method. The proposed method could generate satisfactory dehazing results while effectively suppressing the undesirable artifacts caused by other competing methods.
\begin{figure}[t]
	\centering
	\includegraphics[width=\linewidth]{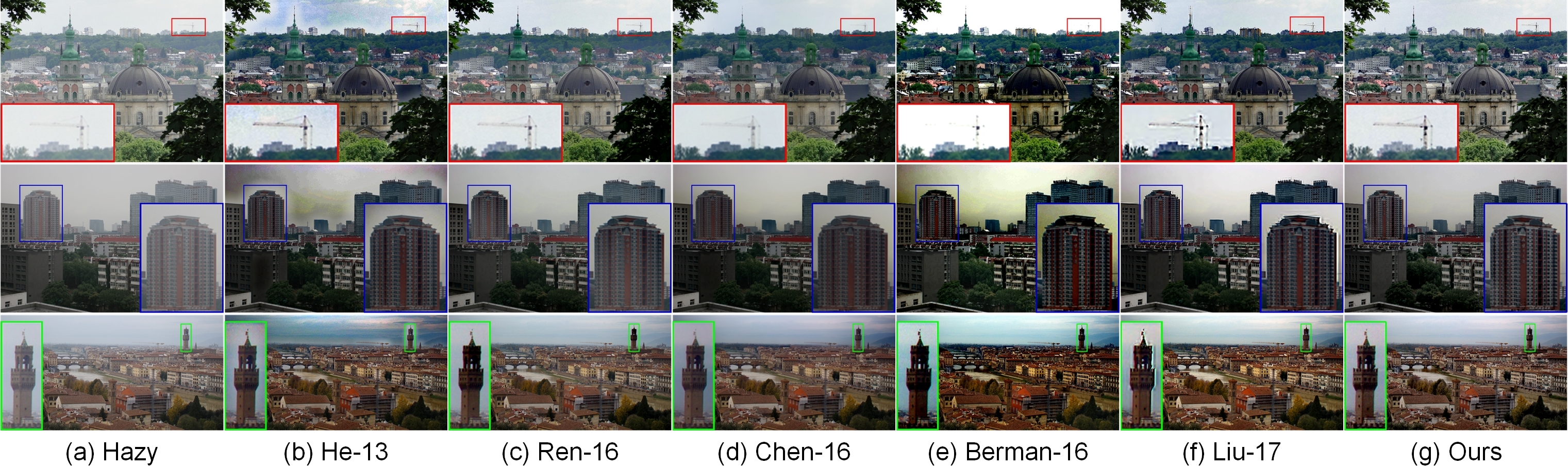}
	\caption{Comparisons of dehazing results on three different realistic images. From left to right: (a) hazy image, dehazed images generated by (b) He-13 \cite{HeSunPAMI2013}, (c) Ren-16 \cite{RenECCV2016}, (d) Chen-16 \cite{ChenECCV2016}, (e) Berman-16 \cite{BermanCVPR2016}, (f) Liu-17 \cite{LiuCVIU2017} and (g) ours.}
	\label{Figure2}
\end{figure}
\subsection{Experiments on Realistic Images}
\begin{figure}[t]
	\centering
	\includegraphics[width=\linewidth]{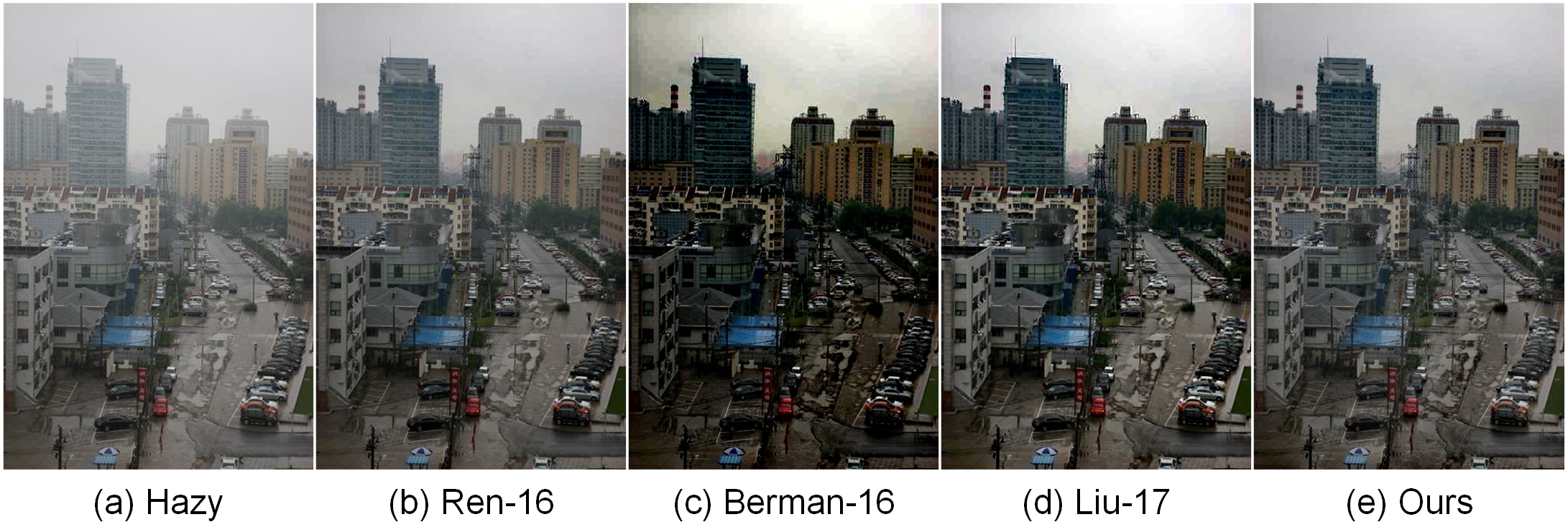}
	\caption{Dehazing results on one cityscape image.}
	\label{Figure3}
\end{figure}

\begin{figure}[t]
	\centering
	\includegraphics[width=\linewidth]{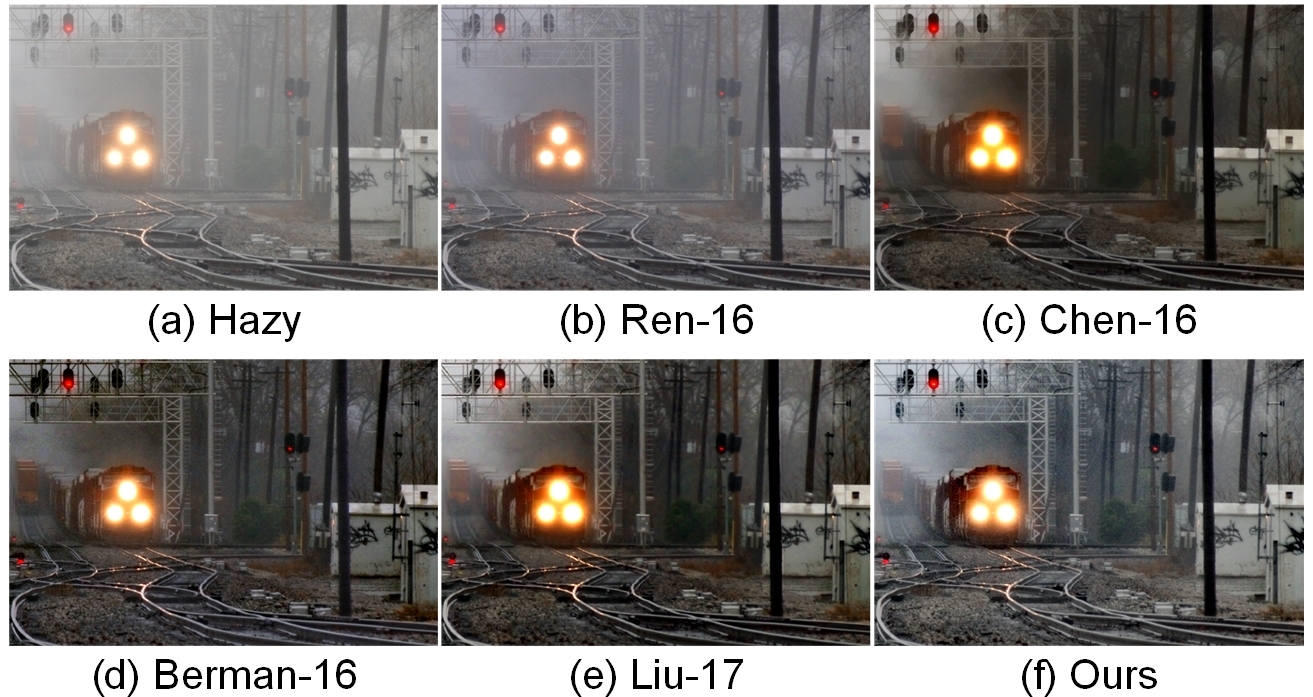}
	\caption{Dehazing results on one train image.}
	\label{Figure4}
\end{figure}

This subsection further implements the comparative dehazing experiments on several realistic degraded images. Fig. \ref{Figure2} visually compares our results to five state-of-the-art image dehazing methods \cite{HeSunPAMI2013, BermanCVPR2016, ChenECCV2016, RenECCV2016, LiuCVIU2017}. We find that He-13 \cite{HeSunPAMI2013} yields the lowest-quality images. Berman-16 \cite{BermanCVPR2016} sometimes suffers from the loss of fine details or color distortion in sky regions. The restored images from Liu-17 \cite{LiuCVIU2017} contain significant ringing artifacts near edges leading to visual quality degradation. In contrast, our dehazing results produced are competitive against Ren-16 \cite{RenECCV2016} and Chen-16 \cite{ChenECCV2016} under these imaging conditions. Dehazing results in Figs. \ref{Figure3} and \ref{Figure4} have also demonstrated our advantages. In particular, the cityscape image in Fig. \ref{Figure3} contains sky region and abundant textures; the train image in Fig. \ref{Figure4} contains headlights which are essentially different from the atmospheric light. It can be visually found that Ren-16 \cite{RenECCV2016} fails to effectively remove the haze. Chen-16 \cite{ChenECCV2016} tends to oversmooth fine image details and degrade image quality. The proposed method is capable of effectively remove haze while preserving fine image details. Its good performance mainly benefits from the weighted fusion-based coarse transmission map estimation and variational regularized transmission map refinement.
\section{Conclusion}
\label{sec:conclusion}
Accurate estimation of transmission map is still a challenging problem of common concern in image dehazing. In this work, the coarse transmission map was first generated by weightedly summing up two different transmission maps, respectively, estimated from foreground and sky regions. To further refine the coarse transmission map, a joint variational regularized model with hybrid constraints was proposed to simultaneously implement transmission map refinement and haze-free image estimation. The resulting nonsmooth optimization problem was effectively solved via an ADMM-based numerical method. Experimental comparisons on both synthetic and realistic images have illustrated our advantages in terms of quantitative and visual quality evaluations. 

\end{document}